\documentclass[10pt,twocolumn,letterpaper]{article}

\usepackage{cvpr}
\usepackage{times}
\usepackage{epsfig}
\usepackage{graphicx}
\usepackage{amsmath}
\usepackage{amssymb}

\usepackage{amsmath,amsfonts,epsfig,graphicx,subfigure,url,multirow,booktabs,collcell}
\usepackage{array,multirow,graphicx}

\usepackage[pagebackref=true,breaklinks=true,letterpaper=true,colorlinks,bookmarks=false]{hyperref}

\cvprfinalcopy 

\def\cvprPaperID{3726} 

\ifcvprfinal\fi
\begin{document}

\title{Joint Network based Attention for Action Recognition}

\author{Yemin Shi$^{1,2}$, Yonghong Tian$^{1,2}$\thanks{Corresponding author: Yonghong Tian (email: yhtian@pku.edu.cn) and Yaowei Wang (yaoweiwang@bit.edu.cn).}, Yaowei Wang$^{3*}$, Tiejun Huang$^{1,2}$\\
$^{1}$ National Engineering Laboratory for Video Technology,
School of EE\&CS, \\
Peking University, Beijing, China\\
$^{2}$ Cooperative Medianet Innovation Center, China\\
$^{3}$ School of Information and Electronics,\\
Beijing Institute of Technology, Beijing, China
}


\maketitle

\begin{abstract}
   By extracting spatial and temporal characteristics in one network, the two-stream ConvNets can achieve the state-of-the-art performance in action recognition. However, such a framework typically suffers from the separately processing of spatial and temporal information between the two standalone streams and is hard to capture long-term temporal dependence of an action. More importantly, it is incapable of finding the salient portions of an action, say, the frames that are the most discriminative to identify the action. To address these problems, a \textbf{j}oint \textbf{n}etwork based \textbf{a}ttention (JNA) is proposed in this study. We find that the fully-connected fusion, branch selection and spatial attention mechanism are totally infeasible for action recognition. Thus in our joint network, the spatial and temporal branches share some information during the training stage. We also introduce an attention mechanism on the temporal domain to capture the long-term dependence meanwhile finding the salient portions. Extensive experiments are conducted on two benchmark datasets, UCF101 and HMDB51. Experimental results show that our method can improve the action recognition performance significantly and achieves the state-of-the-art results on both datasets.
\end{abstract}

\section{Introduction}
Action recognition is the key technique for many visual applications, such as security surveillance, automated driving, home-care nursing, and automatically video tag. Generally speaking, action recognition aims at categorizing the actions or behaviours of one or more persons in a video sequence. Typically, an action can be identified by its spatial (for example, football and piano) or temporal features. In \cite{simonyan2014two}, Simonyan \textit{et al.} proposed the two-stream ConvNets for action recognition. Basically, their method extracts the spatial and temporal characteristics in one framework, and trains the standalone CNNs for two streams separately. However, it is well-known that the spatial and temporal domain are not independent from each other. Naturally, it is beneficial to train the spatial and temporal streams jointly.

In order to make better use of the two-stream framework, several studies made their efforts on fusing the two streams. Simonyan \textit{et al.} \cite{simonyan2014two} tested three ways: training a joint stack of fully-connected layers on top of two streams' features, fusing the softmax scores by averaging and fusing the softmax scores using a linear SVM. They claimed that fusion with fully-connected layers was infeasible due to over-fitting, while SVM-based fusion of softmax scores outperformed the averaging fusion. Wu \textit{et al.} \cite{wu2015modeling} tested more strategies, including SVM-based early fusion, SVM-based late fusion, multiple kernel learning \cite{kloft2011lp}, early fusion with neural networks, late fusion with neural networks, multimodal deep Boltzmann Machines \cite{ngiam2011multimodal,srivastava2012multimodal} and RDNN \cite{wu2014exploring}, and then proposed their Regularized Feature Fusion Network.

However, these two-stream networks often suffer from the so-called ``one-stream-dominating-network'' problem.  Usually, if we use a CNN before fusing the two streams, we should train the CNNs for both spatial and temporal stream at first. Because the temporal stream will converge much slower than the spatial one, training the two streams together will always result in the spatial stream dominating the whole network. It is known that in case of one stream dominating the predictions, little information exchange happens between the two streams.

In action recognition, another important problem is the modeling and utilization of the long-term dependence. It has been proven by many works that better modeling the long-term dependence will improve the performance significantly. Karpathy \textit{et al.} \cite{karpathy2014large} found that a slow fusion in the temporal domain would produce a better result than single frame, late fusion or early fusion. Yue-Hei \textit{et al.} \cite{yue2015beyond} and Donahue \textit{et al.} \cite{donahue2015long} proposed to use recurrent networks by connecting LSTMs to CNNs, and found that RNNs were a better solution than the temporal domain fusion strategy. Shi \textit{et al.} \cite{shi2015learning,shi2016sequential} also introduced their DTD and sDTD to model the dependence on the temporal domain. However, none of them is effective enough for modeling the long-term dependence.

Recently, the attention mechanism \cite{bahdanau2014neural,vinyals2015grammar} was also introduced to action recognition. Sharma \textit{et al.} \cite{sharma2015action} transferred the attention mechanism on the spatial domain to action recognition.  Wu \textit{et al.} \cite{wu2016action} used attention as a regularization to make use of features from different layers in CNN. Unfortunately, no remarkable performance gain was achieved in both works. Obviously, by simply introducing the  attention mechanism, they are totally incapable of finding the salient portions of an action, say, the frames that are most discriminative to identify the action.

In this paper, we propose a \textbf{j}oint \textbf{n}etwork based \textbf{a}ttention (JNA) model which aims at learning the salient portions of actions. Through several exploratory experiments, we find that the fully-connected fusion, branch selection and spatial attention mechanism are totally infeasible for action recognition. So in our joint network, the spatial and temporal branches share some information during the training stage. We also introduce an attention mechanism on the temporal domain to capture the long-term dependence meanwhile finding the salient portions. As a result, our method takes both spatial and temporal stream as input and pulls the most important parts as output. To limit the information exchange between the two streams, their connection is constrained by softmax. Only the crucial information can go through the gate propagating to the lower layers of the other stream during back propagation. Extensive experiments are carried out on two challenging datasets: HMDB51 and UCF101. The results show that the proposed method significantly improves our baseline and achieves the state-of-the-art performance.

The rest of the paper is organized as follows: In section \ref{relatedworks}, we review the related work on action recognition and attention mechanism. We explore the base model before head into the proposed model in section \ref{exploratory_findings}. The proposed joint network based attention model is presented in section \ref{joint_network_based_attention}. We will evaluate our JNA in section \ref{experiments}. Finally, section \ref{conclusion} concludes this paper.

\section{Related works}\label{relatedworks}
\noindent \textbf{Action recognition.} Basically, the action recognition approaches can be categorized into two types by the way of feature extraction: hand-crafted low-level features \cite{dalal2005histograms,dalal2006human} and deep features \cite{wu2015modeling,shi2015learning,wu2014exploring,zha2015exploiting}. The most successful hand-crafted feature is the dense trajectories \cite{wang2011action}, which sample and track dense points from each frame in multiple scales. HOG, HOF and Motion Boundary Histogram (MBH) are also extracted at each point. The combination of these features was shown to further boost the final performance. The improved version of dense trajectories \cite{wang2013action} also takes the camera motion estimation into account and then applies the Fisher vector \cite{perronnin2010improving} to derive the final representation for each video.

In recent years, CNN has achieved state-of-the-art performance on various tasks (e.g. \cite{he2015deep,sutskever2014sequence,szegedy2015going,he2015delving}) and it has been proven that features learnt from CNN are much better than the hand-crafted features. In order to transfer CNNs to video tasks, many models \cite{shi2015learning,zha2015exploiting} have been proposed. Two-stream ConvNets \cite{simonyan2014two} is the most important framework which acts the baseline (with two GRU layers) in many works. Basically, the two-stream ConvNets incorporate spatial and motion networks and pre-train these networks on the large ImageNet \cite{deng2009imagenet} dataset, consequently achieving the state-of-the-art performance.

Unlike these pure deep models, Wang \textit{et al.} \cite{wang2015action} proposed trajectory-pooled deep-convolutional descriptor (TDD), which shares the merits of both hand-crafted features and deeply-learnt features. Shi \textit{et al.} \cite{shi2015learning} proposed the deep trajectory descriptor (DTD) by converting dense trajectories into 2D images and utilizing a CNN to learn features for these images.

\noindent \textbf{Attention mechanism.} The attention mechanism is first introduced to neural machine translation (NMT) by Bahdanau \textit{et al.} \cite{bahdanau2014neural} to automatically learn the alignment between a target word and the relevant parts of the source sentence. Some works did their efforts to apply the attention mechanism to action recognition. Sharma \textit{et al.} \cite{sharma2015action} proposed the spatial attention without modification of the mechanism. Wu \textit{et al.} \cite{wu2016action} proposed a more complicated model and used attention as a regularization. Unfortunately, no remarkable performance gain was achieved by the attention mechanism in both works.

Our joint network based attention (JNA) method will focus on extracting the most important parts in the temporal domain of a video. As a consequence, the improvement can be achieved by fusing the most important frames in two streams.

\section{Exploratory findings}\label{exploratory_findings}
In this section, we will describe the findings from two exploratory experiments including fully-connected fusion and branch selection for fusing two streams. These methods inspired us to propose our JNA method.

\begin{figure}
    \centering
    \subfigure[FC fusion]{
        \includegraphics[width=0.22\textwidth]{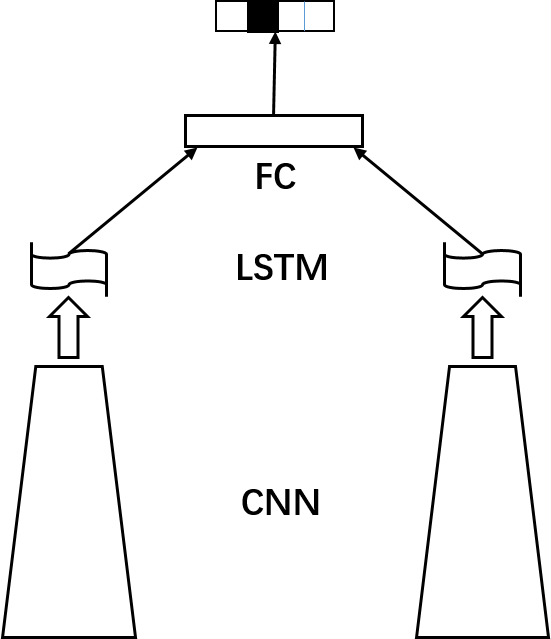}
        \label{figure:fc_fusion}
    }
    \subfigure[Branch selection]{
        \includegraphics[width=0.2\textwidth]{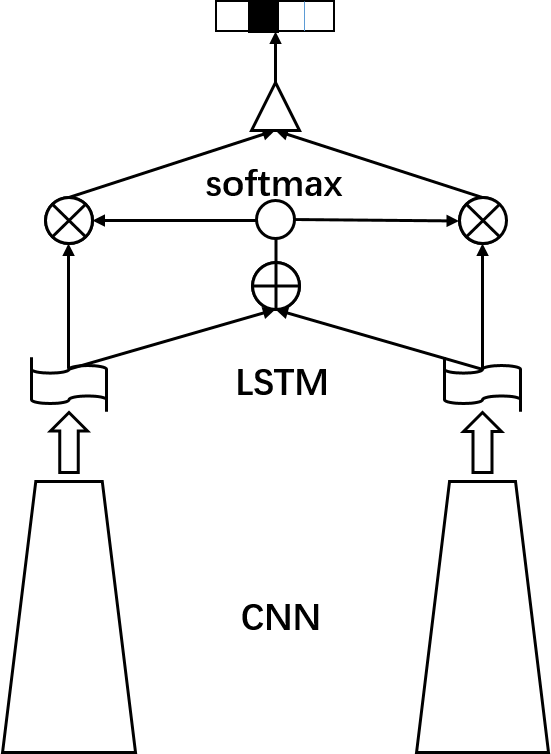}
        \label{figure:branch_selection}
    }
	\caption{The illustration of two exploration fusion methods. The FC fusion uses a fully-connected layer to combine the two streams. The branch selection approach uses an attention model to decide which stream should be selected to produce prediction for the current input. So it works as an automatic weighted averager.}
	\label{figure:fusion}
\end{figure}

\begin{table}
    \centering
    \caption{Comparison of different fusion methods. We use GoogLeNet and two GRU layers, and train the network on UCF101 dataset.}
    \begin{tabular}{c|c|c|c}
        \hline\hline
        Method & Spatial & Temporal & Fusion \\
        \hline
        FC fusion & $80.3\%$ & $80.7\%$ & $81.4\%$ \\
        Branch selection (BS) & - & - & $80.0\%$ \\
        BS with L2 norm & - & - & $83.9\%$ \\
        Average & $80.5\%$ & $82.8 \%$ & $\textbf{90.2\%}$ \\
        \hline\hline
    \end{tabular}
    \label{table:exploration_fusion}
\end{table}

\subsection{Fully-connected fusion}
As shown in Figure \ref{figure:fc_fusion}, a simplest way to fuse two streams is to add another fully-connected (FC) layer on top of them. We first train the two-stream CNNs. Then the FC layer takes the two streams as input and the output is then used to learn a classifier.

The results of FC fusion are listed in Table \ref{table:exploration_fusion}. As reported by \cite{simonyan2014two} and \cite{wu2014exploring}, we find that FC fusion produces much worse results than averaging fusion of softmax scores. The fusion model almost has the same performance as the single spatial or temporal stream. In our opinion, the low accuracy is not only due to the over-fitting problem but also because that one stream dominates the network while the other stream only has a small effect on the final prediction. This assumption is also confirmed by the following branch selection approach.

\subsection{Branch selection}
To begin the discussion, we will first revisit the attention mechanism. In sequence-to-sequence tasks, we have two separate LSTMs (one to encode the sequence of input words $A_i$ and another to produce or decode the output symbols $B_i$). Let $(h_1,h_2,...,h_{T_A})$ denote the hidden states of the encoder while $(d_1,d_2,...,d_{T_B})$ for those of the decoder. To compute the attention vector at each output time $t$ over the input words, we define:
\begin{align*}
& e_i^t=v^T tanh(W_1'h_i+W_2'd_{t-1}) \\
& \alpha_i^t=\frac{exp(e_i^t)}{\sum_{k=1}^{T_A}exp(e_k^t)} \\
& d_t'=\sum_{i=1}^{T_A}\alpha_i^th_i
\end{align*}
where vector $v$ and matrices $W_1', W_2'$ are the parameters of the model. The vector $u^t$ assigns a weight for each encoder hidden state $h_i$, indicating how much attention should be put on $h_i$. These attention weights are normalized by softmax to create the attention mask $a^t$ over the encoder hidden states.

In the two-stream framework, we have to fuse two predictions to get the final result. However, if a video can be predicted correctly by the final prediction, this video is also likely to be correctly predicted by one of the streams. Is it possible to select such a stream for a video?

As illustrated in Figure \ref{figure:branch_selection}, we modify the attention mechanism to take two streams as input and compute attention weights for each stream, as follows:
\begin{align*}
&s=W_1'x_1+W_2'x_2 \\
&e_i=v^T tanh(s+W_3'x_i) \\
&\alpha_i=\frac{exp(e_i)}{\sum_{k=1}^{2}exp(e_k)} \\
&o_i=\sum_{k=1}^2\alpha_ix_i
\end{align*}
where $x_1, x_2$ are the spatial and temporal features respectively. In this model, the output of the branch we are looking at is used twice so that the attention model can generate different weights for two branches. Because that two streams are in different feature space, we apply L2 norm to the inputs:
\begin{align*}
x_1=\lVert x_1 \rVert_2~~~~~~~~x_2=\lVert x_2 \rVert_2
\end{align*}


As shown in Table \ref{table:exploration_fusion},the pure branch selection produces a very similar performance to one stream. Even after applying L2 norm, the performance gain is still small.

\section{Joint network based attention}\label{joint_network_based_attention}
According to previous proposed models, we find that when trying to fuse two streams, we should avoid one stream dominating the output. In this section, we find that applying attention to spatial domain is not effective. However, it works better by joint training two streams with temporal attention model.

\subsection{Spatial attention}

\begin{figure}
	\centerline{\includegraphics[width=0.4\textwidth]{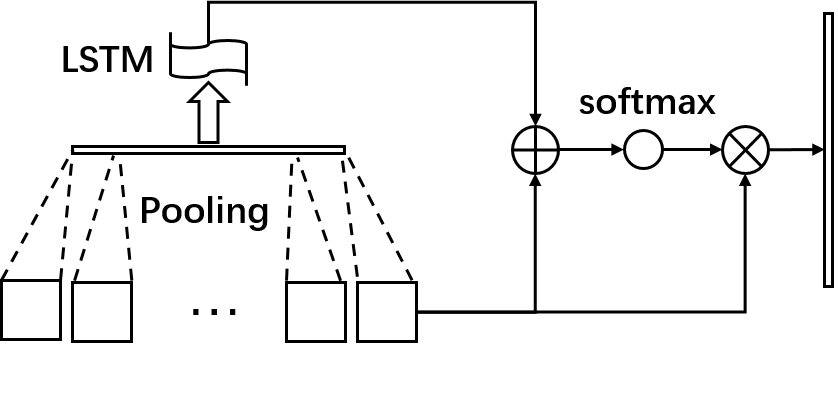}}
	\caption{Illustrating our spatial attention method.}
	\label{figure:spatial_attention}
\end{figure}

In order to make use of the attention mechanism, we also propose the spatial attention model which is designed to replace the last pooling layer.

As shown in Figure \ref{figure:spatial_attention}, we assume the last pooling layer is P, the layer before P is layer C and the feature map size of C is $K\times K$. We add a LSTM layer L after P.~C has $K^2$ positions and $C_{i,j}$ is a vector constructed by the value at $(i,j)$ in all feature maps of C. We compute the attention weight for each position by:
\begin{align*}
&e_{i,j}=v^T tanh(W_1'L+W_2'C_{i,j}) \\
&\alpha_{i,j}=\frac{exp(e_{i,j})}{\sum_{k=1}^{K}\sum_{l=1}^{K}exp(e_{k,l})}
\end{align*}
Then the output of the spatial attention is the weighted average of C by $\alpha$.

Unlike most existing spatial attention approaches, our model employs another LSTM layer to help remember the history information. However, according to Table \ref{table:comparison_with_attentions}, both soft attention or our complicated spatial attention are not able to improve our baselines. These failures prove the infeasibility of spatial attention methods for action recognition.

\subsection{JNA}

\begin{figure*}
	\centerline{\includegraphics[width=0.8\textwidth]{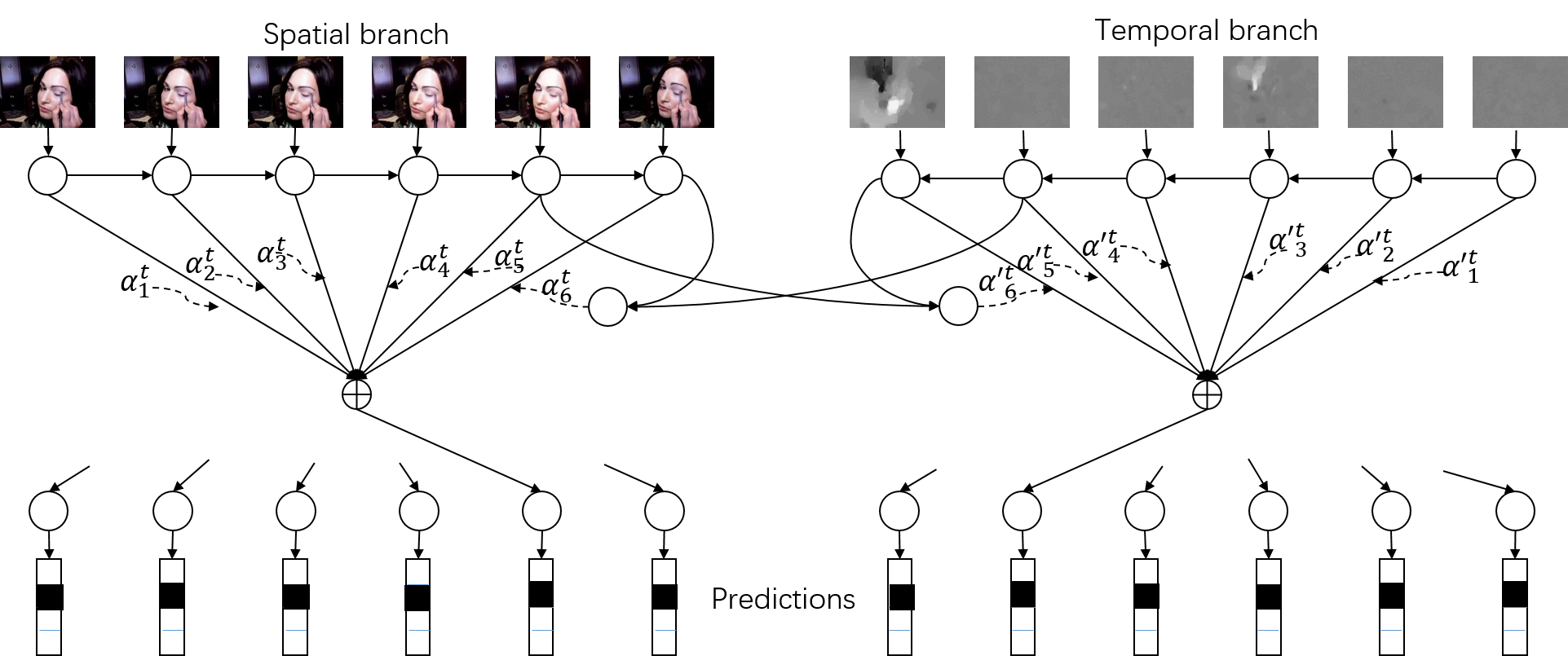}}
	\caption{Illustration of our joint network based attention (JNA).}
	\label{figure:temporal_attention}
\end{figure*}

To overcome the ``one-stream-dominating-the-network'' problem and design a good joint network, we should limit the information exchanging between two streams. This means we must avoid fusion of final features or predictions during training. To satisfy this rule, each stream should have their own softmax and loss layers. On the other hand, two branches should share some layers so that the most crucial information is able to go across two streams during the back propagation. We restrict the information flow by using softmax layer as gate so that information can go through it but only important information can back propagate through this gate.

Our proposed attention model is shown in Figure \ref{figure:temporal_attention}. In JNA, we have two branches, spatial and temporal branches, which have at least one LSTM as the last layers. The outputs of the last LSTM layer in spatial branch are denoted $(h_1,h_2,...,h_T)$ and the outputs of the last LSTM layer of temporal branch are denoted $(g_1,g_2,...,g_T)$. To compute the attention weight for each frame (video frame and optical flow fields), we define:
\begin{align*}
& e_{ij}=v^T tanh(W_1'h_i+W_2'g_j) \\
& f_{ji}=u^T tanh(W_3'g_j+W_4'h_i) \\
& \alpha_{ij}=\frac{exp(e_{ij})}{\sum_{k=1}^T exp(e_{kj})} \\
& \beta_{ji}=\frac{exp(f_{ji})}{\sum_{k=1}^T exp(f_{jk})} \\
& o_j^h=\sum_{i=1}^T \alpha_{ij} h_i \\
& o_i^g=\sum_{j=1}^T \beta_{ji} g_j
\end{align*}
where vector $v, u$ and matrices $W_1', W_2', W_3', W_4'$ are the parameters. Every input feature vector in one branch (denote as A) will be used to compute a group of weights for the other branch (denote as B), and the weights are then used to get weighted average of input feature vectors in B. In this way, the number of output feature vectors in B is equal to the number of input feature vectors in A. The $o_j^h$ and $o_i^g$ are the output of spatial branch and temporal branch respectively and followed by fully-connected layers to learn classifiers.

In this formulation, the information flow is controlled by $\alpha$ and $\beta$. After applying softmax, most of $\alpha$ or $\beta$ are 0 and only the most important inputs have positive weights. This ensures that only the gradients of these important inputs can back propagate to $e$ and $f$, and finally impact both branches. Because that there is no other layer is shared by two branches, $\alpha$ and $\beta$ are the gates to control the information flow which can share across two branches and is called sharing gates.

\section{Experiments}\label{experiments}
This section will first introduce the detail of datasets and their corresponding evaluation schemes. Then, we describe the implementation details of our model. Finally, we report the experimental results and compare JNA with the state-of-the-art methods.

\subsection{Datasets}
To verify the effectiveness of our methods, we conduct experiments on two public datasets: HMDB51 \cite{kuehne2011hmdb} and UCF101 \cite{soomro2012ucf101}.

The HMDB51 dataset is a large collection of realistic videos from various sources, including movies and web videos. It is composed of 6,766 video clips from 51 action categories, with each category containing at least 100 clips. Our experiments follow the original evaluation scheme, but only adopt the first training/testing split. In this split, each action class has 70 clips for training and 30 clips for testing.

The UCF101 dataset contains 13,320 video clips from 101 action classes and there are at least 100 video clips for each class. We tested our model on the first training/testing split in the experiments.

Compared with the very large dataset used for image classification, the dataset for action recognition is relatively smaller. Therefore, we pre-train our model on the ImageNet dataset \cite{deng2009imagenet}. As UCF101 is larger than HMDB51, we also use UCF101 to train our joint model initially, and then transfer the learnt model to HMDB51.

\subsection{Implementation details}
We use the TensorFlow \cite{abadi2016tensorflow} to implement our model and the CNN in every branch is implemented with GoogLeNet \cite{szegedy2015going} structure. We use GRU \cite{cho2014learning} as our LSTM implementation.

The network weights are learnt using the mini-batch stochastic gradient descent with momentum (set to 0.9). The batch size for training CNN is 128 and the batch size for training joint network is 64. When training or testing the joint model, we read 16 frames/flows with a stride of 5 from each video as one sample for the GRU. We resize all input images to $340\times 256$, and then use the fixed-crop strategy \cite{wang2015towards} to crop a $224\times 224$ region from images or their horizontal flip. Because the 16 consecutive samples are needed in the GRU, we also force images from the same video to crop the same region. In the test phase, we sample 4 corners and the center from each image and its horizontal flip, and 25 samples are extracted from each video.

In order to fully train the CNN feature of spatial and temporal branches, we first train two CNN separately. We initialize the CNN with the pre-trained ImageNet model and train CNN classifier like two-stream ConvNets \cite{simonyan2014two}. The trained CNN weights are used to initialize the CNN part of our joint network. Then we train two branches with our JNA method jointly.

For CNN, the learning rate starts from 0.01 and is divided by 10 at iteration $20K$, $30K$ and $35K$, and training is stopped at $40K$ iterations. For joint network, the learning rate is initially set as $10^{-3}$ and divide by 10 at iteration $25K$, $45K$ and $60K$, and training is stopped at $65K$. For the temporal stream, we choose the TVL1 optical flow algorithm \cite{zach2007duality} and the warped TVL1 optical flow field \cite{wang2016temporal}.

In the remainder of the paper, we use spatial stream and temporal stream to indicate the streams in two-stream framework, and each stream is a GoogLeNet followed by two GRU layers. We use spatial branch and temporal branch to indicate the branches in JNA network, and the structure of the two branches is the same as two streams. We use warped spatial branch and warped temporal branch to indicate the two branches in JNA network whose temporal branch input is warped TVL1 optical flow fields, and the JNA is called warped JNA.

\subsection{Evaluation of JNA}
\begin{table}
    \centering
    \caption{Comparison with existing attention methods on HMDB51 and UCF101.}
    \begin{tabular}{c|c|c}
        \hline\hline
        Model & HMDB51 & UCF101 \\
        \hline
        Soft attention \cite{sharma2015action} & $41.3\%$ & - \\
        Multi-branch attention \cite{wu2016action} & $61.7\%$ & $90.6\%$ \\
        \hline
        Spatial attention (SA) & - & $81.95\%$ \\
        SA + pre-train CNN & - & $88.47\%$ \\
        JNA & $\textbf{66.9\%}$ & $\textbf{91.2\%}$ \\
        \hline\hline
    \end{tabular}
    \label{table:comparison_with_attentions}
\end{table}

\noindent \textbf{Comparison with exploratory experiments.} Our JNA is an extension of fully-connected fusion, branch selection and spatial attention methods. We use joint network structure like FC fusion and branch selection while avoid their one-stream-dominating-network problem. JNA learns attention weights on temporal domain while spatial attention learns attention weights on spatial domain. Even though these methods are using similar solution, JNA is the only one which can improve baseline performance and outperform average softmax score fusion. According to Table \ref{table:exploration_fusion} and \ref{table:comparison_with_attentions}, JNA is $1\%$ better than average fusion and other trails are worse than average fusion.

\begin{table}
    \centering
    \caption{Performance of different modules on HMDB51 and UCF101.}
    \begin{tabular}{c|c|c}
        \hline\hline
        Module & HMDB51 & UCF101 \\
        \hline
        Spatial stream & $46.2\%$ & $80.5\%$ \\
        Temporal stream & $50.3\%$ & $82.8\%$ \\
        \hline
        Spatial branch & $50.2\%$ & $81.6\%$ \\
        Temporal branch & $56.9\%$ & $82.5\%$ \\
        Warped spatial branch & $50.3\%$ & $81.2\%$ \\
        Warped temporal branch & $56.9\%$ & $79.1\%$ \\
        \hline
        Two streams & $58.4\%$ & $90.2\%$ \\
        JNA & $\textbf{66.9\%}$ & $\textbf{91.2\%}$ \\
        Warped JNA & $\textbf{66.3\%}$ & $90.0\%$ \\
        \hline\hline
    \end{tabular}
    \label{table:performance_of_modules}
\end{table}

\noindent \textbf{Benefits from JNA.} The performances of different modules are shown in Table \ref{table:performance_of_modules}. JNA can not improve single branch markedly on UCF101, but improves significantly on HMDB51. This may be because that (1) frames in one video of UCF101 do not vary too much and can be well classified by single frame; (2) video lengths of UCF101 are shorter than HMDB51 and selecting important frames for a longer video is much more useful. When considering the final fused model, JNA outperforms two-stream model $1\%$ on UCF101 and $8.5\%$ on HMDB51. The significantly improvement proves that two-stream framework can benefit a lot from JNA.

\noindent \textbf{Comparison with exist video attention.} As shown in Table \ref{table:comparison_with_attentions}, our JNA also achieves much better accuracy than exist attention methods. Although JNA is much simpler than multi-branch attention \cite{wu2016action}, our performance outperforms it a lot, especially on HMDB51 dataset. The inefficiency of soft attention \cite{sharma2015action} also confirms the experiment result of our spatial attention.

\begin{figure*}
	\centerline{\includegraphics[width=0.9\textwidth]{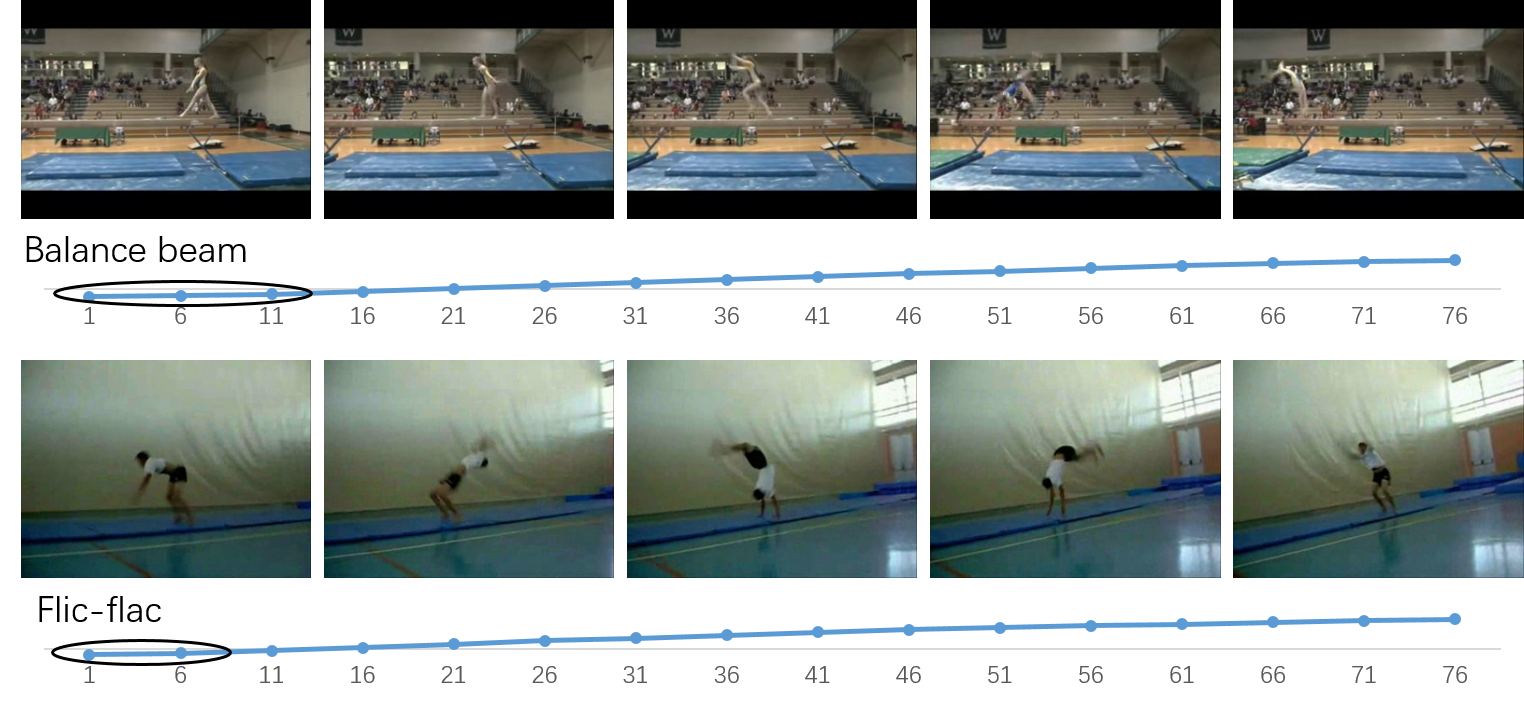}}
	\caption{Visualization of the attention weights (before softmax) in JNA on two videos from UCF101 and HMDB51. The x-axis and y-axis are frame number and attention weight respectively. A higher attention weights means the frame is more salient. The circled parts are the frames which have negative attention weights.}
	\label{figure:attention_curve}
\end{figure*}
\noindent \textbf{Variation of the model's attention.} The visualization curve of the attention weights (before softmax) in JNA are shown in Figure \ref{figure:attention_curve}. Because that our models employ two GRUs, latter frames are able to predict based on history information hence more important than previous frames, and JNA gives bigger weights on the latter frames. This also agrees with our intuition that LSTM like GRU is able to memorize information. However, the variation of attention weights also proves that LSTM can not store all information and methods like JNA can help LSTM improve the performance. The circled points have negative attention weights, which means JNA is able to detect the beginning of action and automatically filter out the irrelevant frames.

\subsection{Comparison with the state-of-the-art methods}

\addtocounter{footnote}{-2}
\begin{table}
    \centering
    \caption{Comparison of JNA to the state-of-the-art methods on HMDB51 and UCF101.}
    \begin{tabular}{c|c|c}
        \hline\hline
        Module & HMDB51 & UCF101 \\
        \hline
        DT+MVSV \cite{cai2014multi} & $55.9\%$ & $83.5\%$ \\
        iDT+HSV \cite{peng2016bag} & $61.1\%$ & $88.0\%$ \\
        Two-stream model \cite{wang2015towards} & $59.4\%$ & $88.0\%$ \\
        F$_{ST}$CN \cite{sun2015human} & $59.1\%$ & $88.1\%$ \\
        TDD+iDT+FV \cite{wang2015action} & $65.9\%$ & $91.5\%$ \\
        Multi-branch attention \cite{wu2016action} & $61.7\%$ & $90.6\%$ \\
        \hline
        JNA & $66.9\%$ & $91.2\%$ \\
        Warped JNA & $66.3\%$ & $90.0\%$ \\
        JNA + Warped JNA & $\textbf{68.8\%}$ & $\textbf{91.5\%}$ \\
        \hline\hline
    \end{tabular}
    \label{table:compare_state_of_the_art}
\end{table}

Table \ref{table:compare_state_of_the_art} compares our results with several state-of-the-art methods on HMDB51 and UCF101 datasets. The performance of JNA significantly surpasses these methods on HMDB51 and outperforms most methods on the UCF101 dataset. The superior performance of our method demonstrates the effectiveness of joint network based attention and justifies the importance of long-term temporal dependence.

\subsection{Conclusion}\label{conclusion}
In this paper, we propose a joint network based attention (JNA) for action recognition, which aims to make two streams benefit from each other and learn to focus on the most discriminative frames of a video. As demonstrated by the experimental results on two challenging datasets, our JNA model can improve the two-stream framework remarkably and achieve state-of-the-art performance. Compared with other methods, JNA is easy to implement while maintaining a similar computational cost.

{\small
\bibliographystyle{ieee}
\bibliography{egbib}
}

\end{document}